\documentclass[11pt,a4paper,copyright]{google}
\usepackage[authoryear,sort&compress,round]{natbib}
\usepackage{tabularx,array}
\usepackage{flafter,placeins}
\usepackage{hyperref}
\usepackage[nameinlink,noabbrev]{cleveref}
\hypersetup{breaklinks=true,colorlinks=true,linkcolor=teal,urlcolor=blue,citecolor=teal,
 pdftitle={Can Language Models Learn to Forecast Stock Prices?},
 pdfsubject={Stock forecasting with selective compaction and Turn-Level PPO}}
\newcommand{\lead}[1]{\noindent\textbf{#1}\ }
\title{Can Language Models Learn to Forecast Stock Prices?}
\author{%
\begin{minipage}{\linewidth}
\centering
{\small\bfseries
Jiacheng Guo\textsuperscript{1},
Suozhi Huang\textsuperscript{1},
Shuzhen Li\textsuperscript{1},
Yunlong Gao\textsuperscript{2},
Zerui Cheng\textsuperscript{1},
Jason Ge,\\[3pt]
Shushu Liang\textsuperscript{3},
Zihao Li\textsuperscript{1},
Hao Lu,
Ming Yin\textsuperscript{1},
Shilong Liu\textsuperscript{1},
Jiashuo Liu,
Xu Kuang\textsuperscript{4},
Mengdi Wang\textsuperscript{1}\\[6pt]}
{\small\normalfont
\textsuperscript{1}Princeton University \quad
\textsuperscript{2}InclusionAI\\[2pt]
\textsuperscript{3}Harvard University \quad
\textsuperscript{4}Stanford University}
\end{minipage}%
}
\date{November 2026}

\begin{document}

\begin{abstract}
Forecasting stock returns is a central task in finance. We study whether language models can learn to perform this task through training. To this end, we build BETA, a chronological sandbox over S\&P 500 stocks in which a model calls tools to retrieve prices, indicators, and news available up to a decision date, and then forecasts the stock’s future return. Evaluating frontier models in this sandbox reveals three systematic weaknesses: their scores decline as the test period approaches the present while a simple momentum baseline stays stable, they predict gains far more often than gains occur, and they predict moves much smaller than those that are realized. We then train models directly in the sandbox. We first fine-tune on filtered, hindsight-generated tool-use trajectories for historical tasks judged predictable from the evidence available at the time, teaching the model how to investigate a stock. We then apply reinforcement learning with rewards based on realized returns. During RL, long investigations frequently exceed the context window. We therefore introduce a selective compaction tool that allows the model to compress chosen portions of its interaction history. Selective compaction changes the context available to subsequent decisions, so policy updates must preserve the context in which each action was generated. Meanwhile, rollouts vary substantially in the number and length of their trainable turns, making a shared group-level baseline less effective. We address these challenges with Context-Aligned Turn-Level PPO, which uses a turn-level critic and computes policy updates using each turn’s actual context. The resulting models, which we call AURA, substantially reduce all three weaknesses. Training improves the 8B model’s score from 24.0 to 47.0 and the 30B model’s from 27.4 to 54.7, with the latter surpassing all evaluated frontier models on our benchmark and achieving its largest margin on the most recent test period. These scores describe tasks selected for alignment between the available evidence and the realized return. Across all 412 candidates in the latest period, the trained 30B model still leads (35.17), while the trained 8B model scores below momentum (25.91 versus 28.33).
\end{abstract}

\maketitle

\begin{figure}[t]
\centering
\includegraphics[width=\linewidth]{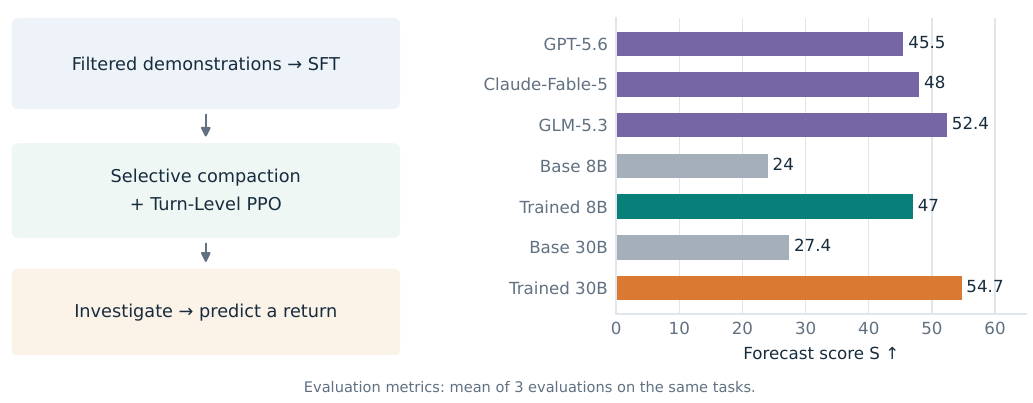}
\caption{\textbf{Learning to forecast in a tool-using stock sandbox.}
Filtered demonstrations initialize the model; selective compaction and Turn-Level PPO support longer investigations. The trained 8B model improves from 24.0 to 47.0, while the trained 30B model reaches 54.7, above the strongest evaluated frontier model at 52.4. Scores are means of three evaluations on the same 396 test tasks.}
\label{fig:overview}
\end{figure}

\section{Introduction}

Finance is an important setting in which language models must reason about an uncertain future. A stock forecast requires more than recalling a fact: an analyst must decide which evidence to inspect, relate observations across time and across companies, and commit to a numerical estimate before the outcome is known. Tool-using language models make this process explicit by collecting evidence through successive interactions \citep{yao2023,schick2023}. The question we study is whether training can improve both that investigation and the resulting forecast.

We build a sandbox on S\&P 500 stocks in which each task specifies a stock, a decision date, and a forecast horizon. The model can retrieve market observations and news available by that date, then submit a future-return prediction. The environment covers 500 stocks, approximately 2.1 million news articles, and eight horizons from one to 126 trading days. We separate training from testing chronologically: the final training label is realized on December 29, 2023, and the first test decision is on May 1, 2025. The 396 test tasks span three successive periods through June 2026, with labels realized by September 30, 2026.

Existing models show three weaknesses in this setting. First, their scores fall substantially across successive test periods, whereas a simple momentum baseline stays near 42. Second, their forecasts are optimistic: they predict gains in 69--76\% of tasks despite an actual positive-return share of 51\%. Third, their predicted moves are too small. The median ratio of predicted to realized absolute return ranges from 0.14 to 0.44 across the evaluated base and frontier models. These observations motivate training a model to investigate evidence and forecast within the sandbox.

We first use supervised fine-tuning (SFT) to teach basic investigation and prediction skills. Demonstrations are generated for historical tasks judged predictable from the evidence available at the time. A teacher can use the historical answer during construction, but demonstrations containing future information or unreasonable forecasting behavior are removed before student training. This yields 3,631 retained demonstrations from 4,420 generated trajectories.

Reinforcement learning introduces two challenges. More extensive investigations can exceed the 32K context window, and rollouts within a group can differ substantially in their number of turns. We equip the agent with selective compaction to replace chosen portions of its interaction history with a summary. We use \emph{Turn-Level PPO} with a turn-level critic and context-aligned policy updates. Each update uses the context observed at that turn and a reward-to-go target. A final forecast reward is combined with a context-budget penalty. This connects the policy update to the context that produced the action while giving each turn its own value estimate.

Training improves forecasting at both scales. At 8B, SFT raises the score from 24.0 to 40.7 and the complete training procedure raises it to 47.0. At 30B, the corresponding scores are 27.4, 46.3, and 54.7. The final 30B model exceeds all evaluated frontier models. Analyses show that the gains extend beyond multiplying base forecasts by a constant, accompany broader evidence gathering, and mitigate all three initial weaknesses. Our contributions are the stock-forecasting sandbox, a training recipe combining selective compaction with Turn-Level PPO, and an analysis of forecasting performance and investigation behavior.

\section{A Sandbox for Stock-Price Forecasting}
\label{sec:sandbox}

\subsection{Task and information access}
A task is a triple $(i,t,H)$: stock $i$, decision date $t$, and horizon $H\in\{1,2,5,10,21,42,63,126\}$ trading days. At the close of the decision date, the model can investigate the stock using tools and then submit one return forecast. The target is the future log return of the daily volume-weighted average price (VWAP):
\begin{equation}
 y_{i,t,H}=\log\!\left(\frac{\mathrm{VWAP}_{i,t+H}}{\mathrm{VWAP}_{i,t}}\right),
 \qquad \widehat{\mathrm{VWAP}}_{i,t+H}=\mathrm{VWAP}_{i,t}\exp(p).
 \label{eq:target}
\end{equation}
Here $p$ is the submitted log-return forecast. Information retrieval is point-in-time: prices, indicators, rankings, market context, and news must be available by $t$. All calls within a task inspect the same historical decision state.

\begin{figure}[t]
\centering
\includegraphics[width=\linewidth]{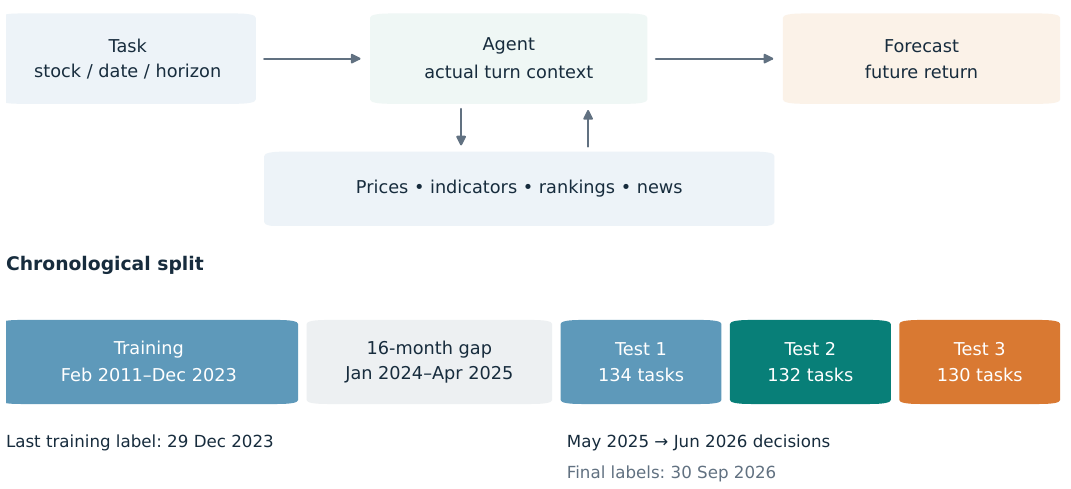}
\caption{\textbf{Sandbox interaction and chronological evaluation.}
The agent repeatedly queries decision-time evidence and submits a return forecast. Training labels end in December 2023; testing starts in May 2025 after a 16-month separation. Test decision windows and label-realization endpoints are different: long-horizon tasks can realize after their decision period ends.}
\label{fig:setting}
\end{figure}

The sandbox contains 11 tools and one terminal action, with a 32K context window. Its data cover 2011 through September 30, 2026. Tools expose individual-stock prices and volume, technical indicators, relative rankings and market context, and news. The news collection contains approximately 2.1 million articles. Selective compaction operates on the interaction history, allowing the model to manage the context used by subsequent turns. Figure~\ref{fig:setting} summarizes the protocol and chronological split.

\subsection{Direction--magnitude score}
\label{sec:score}
We evaluate direction and magnitude jointly:
\begin{equation}
 d(p,y)=\mathbb{1}[\operatorname{sign}(p)=\operatorname{sign}(y),\ p\neq0],\qquad
 m(p,y)=\frac{2\min(|p|,|y|)}{|p|+|y|+\epsilon},\quad \epsilon=10^{-9}.
 \label{eq:score}
\end{equation}
The per-task score is $s(p,y)=d(p,y)m(p,y)$. Wrong-direction and zero forecasts receive zero; a correct-direction forecast earns more when its magnitude is close to the realized move. A prediction of $+3\%$ for a $+5\%$ target receives approximately 75 points, whereas $+0.5\%$ receives approximately 18 and $-1\%$ receives zero.

We evaluate each model three times on the same 396 test tasks and report arithmetic means of the per-evaluation metrics. For each evaluation $r$, directional accuracy $D_r$, magnitude agreement conditional on a correct direction $M_r$, and mean forecast score $S_r$ are defined on a 0--100 scale:
\begin{equation}
 D_r=\frac{100}{N}\sum_j d_{rj},\quad
 M_r=100\frac{\sum_j d_{rj}m_{rj}}{\sum_j d_{rj}},\quad
 S_r=\frac{100}{N}\sum_j d_{rj}m_{rj}=\frac{D_rM_r}{100}.
 \label{eq:aggregate}
\end{equation}
Tables and figures report $D=\frac{1}{3}\sum_{r=1}^{3}D_r$, $M=\frac{1}{3}\sum_{r=1}^{3}M_r$, and $S=\frac{1}{3}\sum_{r=1}^{3}S_r$. Thus, $S$ is the mean of the per-evaluation products, rather than the product of the reported means. Predicted-positive shares and forecast-scale diagnostics are likewise averaged across the three evaluations; task counts and actual positive-return shares describe the shared test set.

We also measure forecast scale using the median ratio $|p|/|y|$ across tasks. This complements $M$, which measures magnitude agreement on correct-direction predictions. All settings use the same direction--magnitude scoring rule \citep{gneiting2011}.

\subsection{Data and chronological separation}
\label{sec:data}
Training uses 3,807 historical tasks across 500 stocks, with decision dates from February 2011 through December 2023. The latest training label is realized on December 29, 2023. There are no training or test decisions in the intervening January 2024--April 2025 gap. The three test periods contain 134, 132, and 130 tasks, respectively. Their positive-return shares are 55\%, 49\%, and 49\%, for a combined share of 51\%. The training share is 54\%.

\begin{table}[htbp]
\centering\small
\begin{tabularx}{\linewidth}{@{}lXXrr@{}}
\toprule
Split & Decision dates & Last label & Stocks & Tasks \\
\midrule
SFT + RL & Feb 2011--Dec 2023 & 29 Dec 2023 & 500 & 3807 \\
Gap & Jan 2024--Apr 2025 & -- & -- & -- \\
Test 1 & May--Sep 2025 & ~1 Apr 2026 & 112 & 134 \\
Test 2 & Oct 2025--Feb 2026 & ~28 Aug 2026 & 109 & 132 \\
Test 3 & Mar--Jun 2026 & 30 Sep 2026 & 108 & 130 \\
\bottomrule
\end{tabularx}

\caption{Decision-date ranges and label endpoints. The first two test label endpoints are approximate. Stocks can appear in multiple test periods.}
\label{tab:splits}
\end{table}

The test horizon mix is closely matched across periods: the short/long counts are 78/56, 77/55, and 77/53. Here short means 1--10 trading days and long means 21--126. Difficulty counts are similarly balanced. Complete counts appear in Appendix~\ref{app:counts}. Training, validation, and the main test evaluation use predictability screening: a task is retained when the evidence available at the decision date supports the subsequently realized move. The realized outcome informs task selection but is hidden from the forecasting model. We separately evaluate all candidates in the final test period.

The start of testing follows the Qwen3 release date of April 29, 2025 \citep{yang2025qwen3}. Other evaluated models have later release dates, listed in Appendix~\ref{app:dates}.

\section{Three Weaknesses of Existing Models}
\label{sec:weaknesses}

\lead{Performance falls in later test periods.}
From the first to the third period, scores decline by roughly 15--45 points across the three frontier models. In the final period, momentum scores 42.48 on the selected tasks, ahead of GPT-5.6 (38.61), Claude-Fable-5 (36.17), and GLM-5.3 (28.09).

\lead{Forecasts lean optimistic.}
GLM-5.3, Claude-Fable-5, and GPT-5.6 predict gains on 70\%, 69.2\%, and 69\% of tasks, respectively, and the base 8B and 30B models on 76\% and 72\%, respectively. The actual positive-return share is 51\%.

\lead{Forecast magnitudes are too small.}
The median absolute forecast-to-outcome ratios are 0.44 for GLM-5.3, 0.37 for Claude-Fable-5, and 0.39 for GPT-5.6. The base 8B and 30B models have still smaller ratios, 0.14 and 0.18. A model may often predict the correct direction while obtaining a modest joint score because its numerical forecast is too conservative. Table~\ref{tab:initial-bias} reports the directional and magnitude diagnostics. Table~\ref{tab:temporal} compares the selected tasks with the full final-period candidate pool.

\section{Training with Selective Compaction and Turn-Level PPO}
\label{sec:method}

\subsection{Supervised initialization from filtered demonstrations}
We first teach investigation and prediction through SFT. Historical tasks are judged predictable from the evidence available at their decision date. An answer-aware teacher generates successful forecasting trajectories for these tasks. The teacher uses realized outcomes to construct demonstrations; the student receives decision-time evidence.

We filter the generated trajectories for future-information leakage and unreasonable forecasting behavior. Of 4,420 generated trajectories, 512 are removed for future information and 277 for unreasonable behavior, retaining 3,631 (82.1\%). The respective removal shares are 11.6\% and 6.3\% of all generated trajectories.

On validation, the base 8B model scores 22.87. Training on all demonstrations from an outcome-blind teacher raises the score to 34.06. Retaining only correct demonstrations raises it to 39.58. With an outcome-aware teacher, the score is 38.12 before filtering and 42.35 after filtering. The filtered recipe exceeds blind-teacher rejection sampling by 2.77 points.

\begin{figure}[t]
\centering
\includegraphics[width=\linewidth]{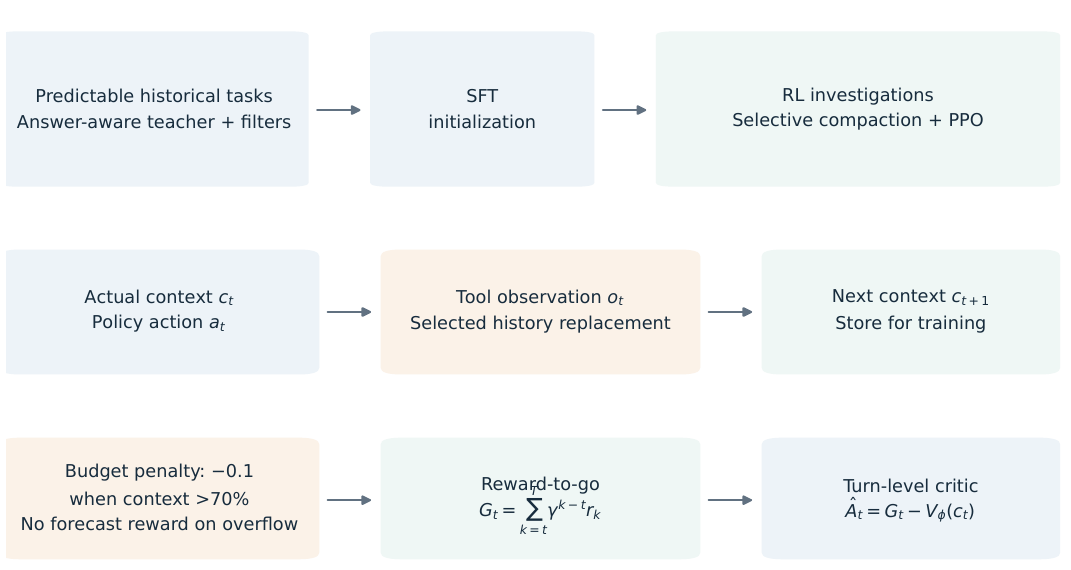}
\caption{\textbf{The training recipe.}
SFT uses filtered historical demonstrations. During RL, selective compaction changes the context seen by later turns. We retain each actual context $c_t$ and use it for policy probabilities and turn-level value estimates. Reward-to-go combines final forecast reward with context-budget penalties. External tool summaries are masked from the actor loss.}
\label{fig:method}
\end{figure}

\subsection{Selective compaction and dynamic context}
\label{sec:compaction}
SFT enables longer investigations, but these investigations create context pressure. Market and indicator tools return approximately 1,100 tokens per call on average, while news returns approximately 2,600. Without compaction, tool-call counts have 10th percentile, median, and 90th percentile values of 9, 17, and 31; the corresponding context-length quantiles are 9K, 24K, and 51K tokens. The fraction exceeding the 32K limit is 2\% for the base model and 36\% after SFT. Among trajectories with at least 25 calls, 91\% exceed the limit.

Context-Folding trains agents to manage working context through branching and summarization \citep{sun2025contextfolding}. In our sandbox, selective compaction lets the model choose a history interval $[l_t,u_t]$. An external tool returns a summary $m_t$, and the chosen interval is replaced while the remainder is preserved. Let $c_t$ denote the actual context observed at turn $t$, $a_t$ the generated action, and $o_t$ its tool observation. We write
\begin{equation}
 c_{t+1}=\operatorname{Update}(c_t,a_t,o_t).
 \label{eq:dynamic}
\end{equation}
For a compaction call, this update includes $\operatorname{Replace}(c_t,[l_t,u_t],m_t)$ together with the action, tool response, and any other new messages. Each sampled turn retains the context that actually produced its action.

The compaction-related reward is a context-budget penalty. At a budget check, a context above 70\% of the maximum receives $-0.1$; exceeding the maximum prevents receipt of the final forecast reward. Let $b_t$ mark an over-threshold budget check associated with turn $t$. The reward supplied to the turn-level return is
\begin{equation}
 r_t=-0.1b_t+\mathbb{1}[t=T]R^{\mathrm{forecast}},
 \qquad R^{\mathrm{forecast}}=s(p,y)
 \label{eq:reward}
\end{equation}
for a valid completed forecast, with no forecast reward on context overflow. This penalty encourages timely context management; predictive usefulness of retained information is reflected in the final forecast score.

\subsection{Reward-to-go and a turn-level critic}
\label{sec:ppo}
Rollouts differ in their number of turns and in the information gathered before prediction. A group-level baseline shares one reference across these different trajectories. Turn-level critics have proved effective for long-horizon game agents in Odysseus \citep{shi2026odysseus}. We estimate future return from each forecasting turn's actual context:
\begin{equation}
 G_t=\sum_{k=t}^{T}\gamma^{k-t}r_k,\qquad
 V_\phi(c_t)\approx\mathbb{E}_{\pi}[G_t\mid c_t],\qquad
 \hat A_t=G_t-V_\phi(c_t).
 \label{eq:advantage}
\end{equation}
The critic is fitted to these turn-level targets using
\begin{equation}
 \mathcal L_V=\frac{1}{T}\sum_{t=1}^{T}\bigl(V_\phi(c_t)-G_t\bigr)^2.
 \label{eq:critic}
\end{equation}
Reward-to-go assigns only current and future rewards to the current action. The critic supplies a context-dependent baseline at every turn. We use these estimates within the clipped PPO update \citep{schulman2017}.

A turn can generate multiple policy tokens. Let $a_{t,k}$ denote token $k$ within turn $t$, with $n_t$ policy tokens. The importance ratio uses the same stored context under both policies:
\begin{equation}
 \rho_{t,k}(\theta)=
 \frac{\pi_\theta(a_{t,k}\mid c_t,a_{t,<k})}
 {\pi_{\mathrm{old}}(a_{t,k}\mid c_t,a_{t,<k})}.
 \label{eq:ratio}
\end{equation}
We preserve $c_t$, including prior compaction results, for both policy evaluation and critic fitting. This keeps each update aligned with the context that produced the action.

We express the actor objective with equal turn weighting and a token mean within each turn:
\begin{equation}
 \mathcal L_{\mathrm{actor}}=-\frac{1}{T}\sum_{t=1}^{T}\frac{1}{n_t}
 \sum_{k=1}^{n_t}\min\!\left(
 \rho_{t,k}\hat A_t,
 \operatorname{clip}(\rho_{t,k},1-\epsilon_{\mathrm{PPO}},1+\epsilon_{\mathrm{PPO}})\hat A_t
 \right).
 \label{eq:ppo}
\end{equation}
This normalization gives each turn equal weight. The model-generated choice of compaction interval and tool parameters belongs to the action. A summary produced by the external compaction tool belongs to the observation and is excluded from the actor-token loss.

\section{Experimental Results}
\label{sec:results}

\subsection{Main forecasting results}
All main test results average three evaluations on the same 396 tasks from the three test periods. Table~\ref{tab:main} reports $D$, $M$, and $S$ for the momentum baseline, frontier models, and base/SFT/full-training checkpoints at 8B and 30B. At 8B, the base, SFT, and RL checkpoints score 23.96, 40.71, and 47.04. This exceeds GPT-5.6 at 45.5 and approaches Claude-Fable-5 at 48.0, while remaining below GLM-5.3 at 52.4. At 30B, the corresponding scores are 27.43, 46.28, and 54.69. The final checkpoint leads GLM-5.3 (52.38) by 2.31 points.

\begin{table}[htbp]
\centering\small
\begin{tabularx}{\linewidth}{@{}Xrrr@{}}
\toprule
Setting & D & M & S \\
\midrule
Momentum & 60.0 & 70.0 & 42.0 \\
GPT-5.6 & 78.2 & 58.2 & 45.5 \\
Claude-Fable-5 & 79.4 & 60.4 & 48.0 \\
GLM-5.3 & 81.6 & 64.2 & 52.4 \\
Base 8B & 65.1 & 36.8 & 24.0 \\
SFT 8B & 68.4 & 59.5 & 40.7 \\
Trained 8B & 72.1 & 65.2 & 47.0 \\
Base 30B & 67.9 & 40.3 & 27.4 \\
SFT 30B & 72.4 & 64.0 & 46.3 \\
Trained 30B & 78.3 & 69.8 & 54.7 \\
\bottomrule
\end{tabularx}

\caption{\textbf{Main results on all 396 test tasks.} $D$ is direction accuracy, $M$ is magnitude agreement conditional on a correct direction, and $S$ is their joint score. Values are means of three evaluations on the same tasks, on a 0--100 scale. Higher is better.}
\label{tab:main}
\end{table}

SFT accounts for a substantial part of the gain, reaching 40.7 at 8B and 46.3 at 30B. RL adds another 6.3 and 8.4 points, respectively. The trained 30B model's directional accuracy is 78.3, below GLM-5.3's 81.6, but its magnitude agreement is higher (69.8 versus 64.2), producing the stronger joint score.

\begin{figure}[t]
\centering
\includegraphics[width=\linewidth]{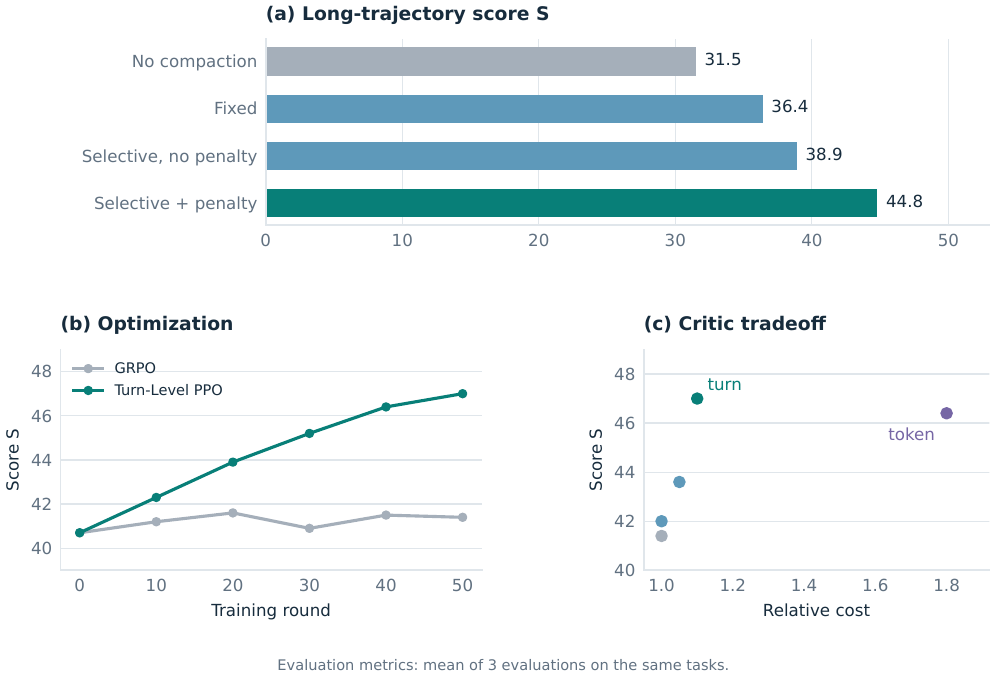}
\caption{\textbf{Evidence for the training design at 8B.}
(a) Selective compaction with the context-budget penalty has the strongest long-trajectory score. (b) Turn-Level PPO improves from 40.7 to 47.0 across the six checkpoints, while GRPO ends at 41.4. (c) The turn-level critic obtains 47.0 at 1.1 times the reference training cost; a token-level critic obtains 46.4 at 1.8 times the cost.}
\label{fig:training}
\end{figure}

\subsection{Compaction enables longer investigations}
Without compaction, 36\% of trajectories exceed the context window, the long-trajectory score is 31.5, and the overall score is 42.1. Fixed compaction summarizes all history once context reaches 24K tokens. It lowers overflow to 1\% and raises the overall score to 43.2, but its long-trajectory score remains 36.4.

Selective compaction without the budget penalty yields 13\% overflow, a long-trajectory score of 38.9, and an overall score of 43.8. Adding the penalty gives 4\% overflow, a long-trajectory score of 44.8, and the full 47.0 overall score. Fixed compaction has the lowest overflow, but selective compaction with the penalty has the best forecasting results. Median context lengths and compaction counts appear in Table~\ref{tab:compaction}.

\subsection{Turn-level values improve optimization}
Starting from the 40.7 SFT checkpoint, GRPO stays between 40.7 and 41.6 and ends at 41.4. Turn-Level PPO rises through 42.3, 43.9, 45.2, 46.4, and 47.0 at the subsequent checkpoints.

The critic ablation separates several alternatives. PPO without a learned critic reaches 42.0. A single value per trajectory reaches 43.6 at 1.05 times the reference cost. A token-level critic reaches 46.4 at 1.8 times the cost, while the turn-level critic reaches 47.0 at 1.1 times the cost. Critic explained variance increases from 0.08 in the first trajectory third to 0.21 in the middle and 0.37 in the last. Table~\ref{tab:critic} and Appendix~\ref{app:optimization} provide the full values.

\paragraph{Validation after 100 RL steps.} Starting from the 42.35 SFT checkpoint, context-aligned turn-level PPO reaches 48.07. Removing the context-budget penalty lowers the score to 45.63. Fixed compaction yields 44.12 and disabling compaction yields 43.47. With a token-level critic, PPO scores 46.58, while GRPO scores 42.86.

\subsection{Component ablations and model scale}
\begin{figure}[t]
\centering
\includegraphics[width=\linewidth]{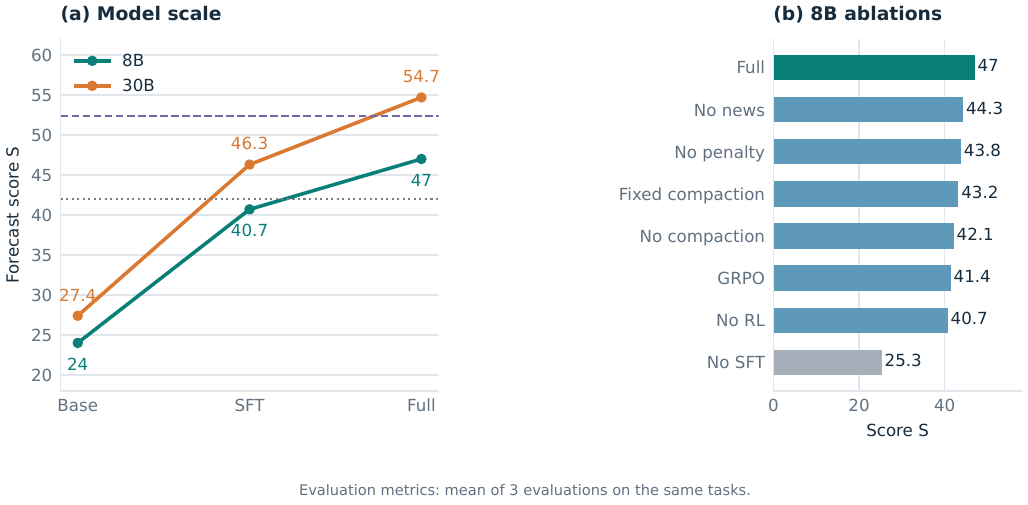}
\caption{\textbf{Scale and components.}
Both model sizes improve through Base, SFT, and full training. The 30B endpoint exceeds the highest evaluated frontier score. In the 8B ablations, every removed or replaced component lowers the score; SFT initialization has the largest observed ablation gap.}
\label{fig:results}
\end{figure}

Figure~\ref{fig:results} summarizes scale and component ablations. Removing news lowers the 8B score from 47.0 to 44.3. Removing the compaction-related penalty lowers it to 43.8, using fixed compaction to 43.2, and removing compaction to 42.1. Replacing PPO with GRPO yields 41.4; removing RL gives the 40.7 SFT checkpoint. Removing SFT gives 25.3, emphasizing the importance of supervised initialization for this training recipe.

\section{What Changes After Training?}
\label{sec:analysis}

\subsection{Magnitude agreement explains most of the numerical gain}
To describe changes in the product of the mean direction and magnitude scores, define $\widetilde S_i=D_iM_i/100$ for checkpoint $i$. We use the symmetric identity
\begin{align}
 \Delta S_D&=\frac{M_0+M_1}{200}(D_1-D_0),&
 \Delta S_M&=\frac{D_0+D_1}{200}(M_1-M_0),\\
 \widetilde S_1-\widetilde S_0&=\Delta S_D+\Delta S_M.
 \label{eq:decomposition}
\end{align}
For 8B, the direction-associated contribution is 3.57 points and the magnitude-associated contribution is 19.48 points, or 84.5\% of the recomposed gain. For 30B, the respective contributions are 5.73 and 21.56 points, with magnitude accounting for 79.0\%. The decomposition uses the displayed mean $D/M$ values. Its recomposed gain can differ from the mean $S$ gain because of across-evaluation covariance and rounding.

\subsection{The gains are not explained by a constant rescaling}
Multiplying base 8B forecasts by a positive constant preserves their predicted signs. With scaling factors fitted on validation and applied to the test set, the best global scale raises $S$ from 24.0 to 31.6; fitting a separate scale per horizon reaches 34.2. Both remain below the trained model's 47.0. Magnitude rank correlation is 0.06 for the base model, remains 0.06 under global rescaling, reaches 0.11 under per-horizon rescaling, and is 0.31 after training.

As another diagnostic, shuffling trained predictions among tasks with the same horizon reduces $S$ to 28.3 and rank correlation to 0.02. This destroys the association between each task and its own forecast while retaining the forecasts available within each horizon. The deterioration indicates that the trained predictions contain task-specific information beyond their overall scale. Table~\ref{tab:controls} includes all direction and magnitude values.

\begin{figure}[t]
\centering
\includegraphics[width=\linewidth]{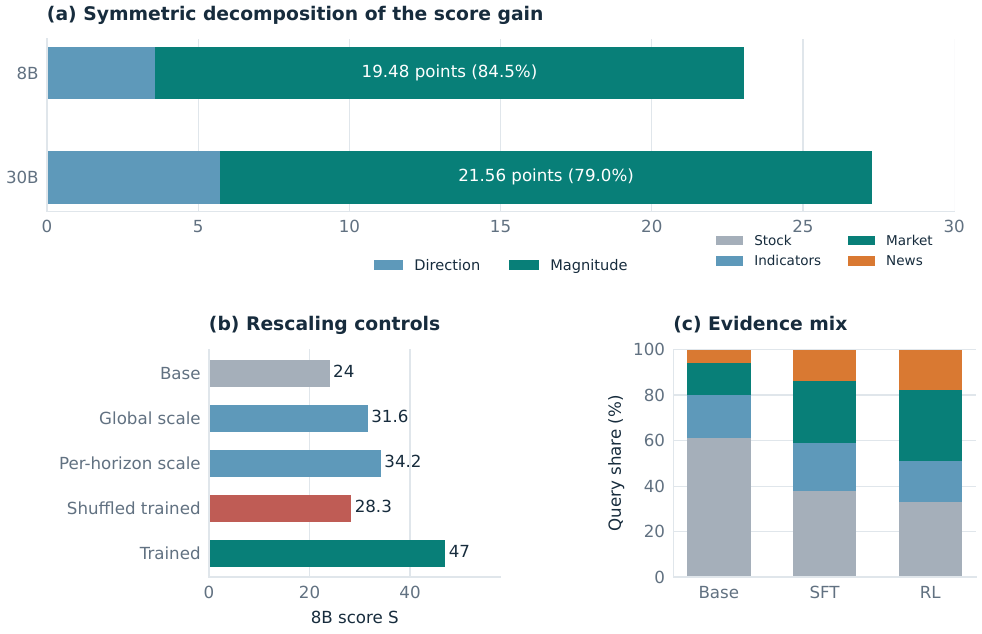}
\caption{\textbf{What changes with training.}
(a) Symmetric decomposition attributes most of the numerical score increase to magnitude agreement. (b) Constant-rescaling controls do not reach the trained 8B score, and within-horizon shuffling removes much of its advantage. (c) Training broadens the evidence mix toward ranking, market context, and news.}
\label{fig:learning}
\end{figure}

\subsection{Investigation becomes broader}
The median tool-call count changes from 4 for the base 8B model to 19 after SFT and 17 after RL. Individual-stock queries fall from 61\% to 38\% to 33\% of queries. Ranking and market-context queries rise from 14\% to 27\% to 31\%, and news queries from 6\% to 14\% to 18\%. Technical-indicator shares are 19\%, 21\%, and 18\%. Compactions per trajectory rise from zero to 1.1 after SFT and 1.6 after RL. SFT thus substantially expands investigation; RL preserves broad evidence gathering with a slightly lower median call count and more compaction.

Grouping individual-stock queries with technical indicators, their combined share falls from 80\% before training to 59\% after SFT and 51\% after RL. The number of calls also varies with difficulty. After SFT, the three difficulty groups span 16--21 calls per task. After RL, easy tasks receive 11 calls, medium tasks 17, and hard tasks 24.

\paragraph{Retrieving versus using evidence.} We give the predictor the retrieved tool outputs with the investigator's reasoning removed. The base 8B model scores 23.12 on its own retrievals and 31.23 on the trained model's retrievals, an 8.11-point gain. Holding those trained retrievals fixed, switching to the trained predictor raises the score by another 13.34 points, to 44.57. The larger improvement in this comparison comes from interpreting the same evidence. End-to-end investigation with the trained model's own reasoning scores 47.04.

\subsection{All three weaknesses are mitigated}
The trained models lose roughly 2--3 points between the first and final test periods. Their predicted positive-return shares fall to 55\% and 54\% at 8B and 30B, respectively, closer to the observed 51\%. Their median absolute forecast-to-outcome ratios rise to 0.79 and 0.86.

\subsection{Horizon and difficulty}
The trained 8B model scores 49.3 on short horizons and 43.7 on long horizons, while the trained 30B model scores 56.4 and 52.3. GLM-5.3 scores 48.9 and 57.4. GLM-5.3 leads on long-horizon and easy tasks, while the trained 30B model leads overall. On hard tasks, the trained 8B and 30B models obtain 29.8 and 37.1, compared with 27.4 for GLM-5.3. Complete subgroup results appear in Table~\ref{tab:subgroups}.

\subsection{Evaluation without task selection}
Of the 412 candidate tasks in March--June 2026, 130 pass the predictability screen. On all candidates, the trained 30B score drops from 54.23 to 35.17 but remains the highest among the six systems (Table~\ref{tab:temporal}). The 8B score drops from 46.52 to 25.91, below both momentum (28.33) and GPT-5.6 (26.55). The ranking on selected tasks therefore transfers only partly to the wider candidate pool.

\section{Related Work}
Language-model forecasting includes approaches that gather evidence for event forecasts \citep{halawi2024} and learn from realized outcomes \citep{turtel2026}. Financial studies investigate return predictability from language-model judgments \citep{lopezlira2023} and the risk of look-ahead bias in historical evaluation \citep{glasserman2023}. Our setting combines point-in-time evidence gathering with a continuous return target and chronological evaluation of a trained tool-using policy.

Time-series approaches use language-model representations or specialized forecasting models \citep{gruver2023,jin2024,ansari2024}. Our model instead chooses which observations to retrieve and when to end its investigation. Domain adaptation for finance, including BloombergGPT and FinGPT \citep{wu2023bloomberg,yang2023fingpt}, addresses a broader set of financial language tasks. Financial agents such as FinMem \citep{yu2023finmem} use memory in sequential decision making. Here, selective compaction is a policy-controlled operation on interaction history, and policy updates must preserve the context produced by that operation.

PPO provides the clipped policy objective underlying our training \citep{schulman2017}. Outcome-based post-training has also improved language-model reasoning \citep{deepseek2025}. Odysseus trains long-horizon game agents with PPO and a lightweight turn-level critic \citep{shi2026odysseus}. Context-Folding uses branch and return actions to reduce active context, and FoldGRPO trains this behavior using folded contexts and process rewards \citep{sun2025contextfolding}. Our forecasting agent uses selective history compaction, context-aligned PPO updates, and a turn-level critic with forecast and context-budget rewards.

\section{Limitations}
The main evaluation covers 396 selected tasks. Screening favors cases in which contemporaneous evidence aligns with the realized move. Evaluating all candidates in the latest period lowers every score (Table~\ref{tab:temporal}). Neither evaluation measures portfolio returns after trading costs or risk. A portfolio-level evaluation across all eligible decisions could measure returns after transaction costs and quantify drawdowns. The stock universe is limited to the S\&P 500, and the available evidence depends on the coverage of the sandbox news and market-data tools.

\section{Conclusion}
Language models can learn to forecast stock prices more effectively through training in a tool-using sandbox. Existing models show declining later-period performance, an optimistic directional bias, and underpredicted move sizes. Filtered supervised demonstrations, selective compaction, and context-aligned Turn-Level PPO improve forecasting at both 8B and 30B and substantially mitigate all three weaknesses. The 8B model rises from 24.0 to 47.0, narrowing the gap to frontier models, while the 30B model reaches 54.7 and exceeds the evaluated frontier systems. The gains accompany broader evidence gathering and task-specific forecast information beyond a simple change of output scale.

\FloatBarrier
\begingroup
\raggedright
\bibliographystyle{abbrvnat}
\bibliography{refs}
\endgroup

\clearpage
\appendix
\section{Dataset Composition and Calendar Details}
\label{app:counts}
\begin{table}[htbp]\centering\small
\begin{tabularx}{\linewidth}{@{}Xrrrrrrrrr@{}}
\toprule
Split & 1 & 2 & 5 & 10 & 21 & 42 & 63 & 126 & Total \\
\midrule
Train & 412 & 398 & 561 & 604 & 547 & 489 & 431 & 365 & 3807 \\
Test 1 & 16 & 15 & 22 & 25 & 20 & 16 & 12 & 8 & 134 \\
Test 2 & 16 & 15 & 22 & 24 & 20 & 16 & 12 & 7 & 132 \\
Test 3 & 16 & 15 & 22 & 24 & 19 & 16 & 11 & 7 & 130 \\
Test total & 48 & 45 & 66 & 73 & 59 & 48 & 35 & 22 & 396 \\
\bottomrule
\end{tabularx}

\caption{Task counts by horizon in trading days. All 396 test tasks contribute to the reported main score.}
\label{tab:horizons}\end{table}
\begin{table}[htbp]\centering\small
\begin{tabularx}{\linewidth}{@{}Xrrrrr@{}}
\toprule
Split & Short & Long & Easy & Medium & Hard \\
\midrule
Test 1 & 78 & 56 & 46 & 53 & 35 \\
Test 2 & 77 & 55 & 45 & 53 & 34 \\
Test 3 & 77 & 53 & 45 & 52 & 33 \\
Total & 232 & 164 & 136 & 158 & 102 \\
\bottomrule
\end{tabularx}

\caption{Short/long horizon and difficulty counts. Short horizons are 1--10 trading days; long horizons are 21--126.}
\label{tab:counts}\end{table}
Training and test decisions are separated by a calendar gap, and every training label is realized before the first test decision. Long test horizons explain why some labels realize well after their decision period. Stocks can appear in multiple test periods.

\subsection{Release milestones}
\label{app:dates}
\begin{table}[htbp]\centering\small
\begin{tabularx}{\linewidth}{@{}XX@{}}
\toprule
Event & Date \\
\midrule
Last training label & 29 December 2023 \\
Qwen3 release & 29 April 2025 \\
First test decision & 1 May 2025 \\
Claude-Fable-5 release & 9 June 2026 \\
Last test decision & 30 June 2026 \\
GPT-5.6 release & 9 July 2026 \\
GLM-5.3 release & 14 August 2026 \\
Last test label & 30 September 2026 \\
\bottomrule
\end{tabularx}

\caption{Training, testing, and model-release dates.}
\label{tab:milestones}\end{table}

\FloatBarrier
\section{Final-Period Task Selection and Forecast Bias}
\begin{table}[htbp]\centering\small
\begin{tabular}{lrr}
\toprule
Model & Selected (130) & All candidates (412) \\
\midrule
Trained 30B & 54.23 & 35.17 \\
Momentum & 42.48 & 28.33 \\
GPT-5.6 & 38.61 & 26.55 \\
Trained 8B & 46.52 & 25.91 \\
Claude-Fable-5 & 36.17 & 25.25 \\
GLM-5.3 & 28.09 & 19.27 \\
\bottomrule
\end{tabular}
\caption{Effect of task selection in the final test period (March--June 2026). The full candidate set includes the 130 selected tasks.}
\label{tab:temporal}\end{table}
\begin{table}[htbp]\centering\small
\begin{tabularx}{\linewidth}{@{}Xrr@{}}
\toprule
Model & Predicted up & Median magnitude ratio \\
\midrule
GLM-5.3 & 70\% & 0.44 \\
Claude-Fable-5 & 69.2\% & 0.37 \\
GPT-5.6 & 69\% & 0.39 \\
Base 8B & 76\% & 0.14 \\
Base 30B & 72\% & 0.18 \\
\bottomrule
\end{tabularx}

\caption{Initial directional bias and forecast scale, averaged over three evaluations. Actual positive returns account for 51\% of test tasks.}
\label{tab:initial-bias}\end{table}
\begin{table}[htbp]\centering\small
\begin{tabularx}{\linewidth}{@{}Xrr@{}}
\toprule
Model & Predicted up & Median magnitude ratio \\
\midrule
GLM-5.3 & 70\% & 0.44 \\
Base 8B & 76\% & 0.14 \\
Trained 8B & 55\% & 0.79 \\
Base 30B & 72\% & 0.18 \\
Trained 30B & 54\% & 0.86 \\
\bottomrule
\end{tabularx}

\caption{Directional bias and forecast scale after training, with base models and GLM-5.3 for comparison.}
\label{tab:post-bias}\end{table}

\FloatBarrier
\section{Demonstration and SFT Ablations}
\begin{table}[htbp]\centering\small
\begin{tabularx}{\linewidth}{@{}Xrr@{}}
\toprule
Step & Trajectories & Share of generated \\
\midrule
Teacher-generated & 4420 & 100.0\% \\
Removed: future information & 512 & 11.6\% \\
Removed: unreasonable forecasting & 277 & 6.3\% \\
Retained for SFT & 3631 & 82.1\% \\
\bottomrule
\end{tabularx}

\caption{Teacher-trajectory filtering. Percentages use all generated trajectories as the denominator.}
\label{tab:filtering}\end{table}
\begin{table}[htbp]\centering\small
\begin{tabularx}{\linewidth}{@{}Xr@{}}
\toprule
Setting & S \\
\midrule
No SFT & 24.0 \\
Teacher without answers & 32.1 \\
No predictability screening & 34.4 \\
Answer-aware, unfiltered & 36.5 \\
Answer-aware, filtered & 40.7 \\
\bottomrule
\end{tabularx}

\caption{SFT-data ablations at 8B.}
\label{tab:sft}\end{table}
The teacher uses historical answers to construct demonstrations. Filtering removes future information from student-visible trajectories. The task-screening ablation removes predictability-based selection of training tasks.

\FloatBarrier
\section{Context Management and Optimization}
\label{app:optimization}
\begin{table}[htbp]\centering\small
\begin{tabularx}{\linewidth}{@{}Xrrr@{}}
\toprule
Statistic & P10 & Median & P90 \\
\midrule
Tool calls & 9 & 17 & 31 \\
Context tokens & 9K & 24K & 51K \\
\bottomrule
\end{tabularx}

\caption{Tool-call and context-length quantiles without compaction.}
\label{tab:context}\end{table}
\begin{table}[htbp]\centering\small
\begin{tabularx}{\linewidth}{@{}Xrrrrr@{}}
\toprule
Setting & Overflow & Context & Comp. & Long S & S \\
\midrule
No compaction & 36\% & 24K & 0.0 & 31.5 & 42.1 \\
Fixed compaction & 1\% & 10K & 1.3 & 36.4 & 43.2 \\
Selective, no penalty & 13\% & 17K & 0.6 & 38.9 & 43.8 \\
Selective + budget penalty & 4\% & 12K & 1.6 & 44.8 & 47.0 \\
\bottomrule
\end{tabularx}

\caption{Compaction comparisons at 8B. Context is the median length in thousands of tokens; Comp. is mean compactions per trajectory. Long S is the score on trajectories with at least 25 tool calls. Fixed compaction summarizes all history at 24K tokens.}
\label{tab:compaction}\end{table}
\begin{table}[htbp]\centering\small
\begin{tabularx}{\linewidth}{@{}Xrr@{}}
\toprule
Round & GRPO & Turn-Level PPO \\
\midrule
0 & 40.7 & 40.7 \\
10 & 41.2 & 42.3 \\
20 & 41.6 & 43.9 \\
30 & 40.9 & 45.2 \\
40 & 41.5 & 46.4 \\
50 & 41.4 & 47.0 \\
\bottomrule
\end{tabularx}

\caption{Forecast scores at six training checkpoints.}
\label{tab:curve}\end{table}
\begin{table}[htbp]\centering\small
\begin{tabularx}{\linewidth}{@{}Xrr@{}}
\toprule
Setting & Final S & Relative cost \\
\midrule
GRPO: group baseline & 41.4 & 1x \\
PPO: no learned critic & 42.0 & 1x \\
PPO: trajectory critic & 43.6 & 1.05x \\
PPO: token critic & 46.4 & 1.8x \\
PPO: turn critic & 47.0 & 1.1x \\
\bottomrule
\end{tabularx}

\caption{Critic granularity, final score and relative training cost. GRPO and PPO without a learned critic each have reference cost 1.0.}
\label{tab:critic}\end{table}
Critic explained variance is 0.08, 0.21, and 0.37 in the first, middle, and last trajectory thirds. The actor objective in Equation~\ref{eq:ppo} averages tokens within each turn, then averages turns; external tool observations are excluded from the token loss. The same stored turn context is used for old-policy likelihoods, new-policy likelihoods, and critic estimates.

\FloatBarrier
\section{Additional Result and Behavior Tables}
\begin{table}[htbp]\centering\small
\begin{tabularx}{\linewidth}{@{}Xrr@{}}
\toprule
Setting & S & Drop from full \\
\midrule
Full training & 47.0 & 0.0 \\
Without news & 44.3 & 2.7 \\
Without context penalty & 43.8 & 3.2 \\
Fixed instead of selective & 43.2 & 3.8 \\
Without compaction & 42.1 & 4.9 \\
GRPO instead of PPO & 41.4 & 5.6 \\
Without RL & 40.7 & 6.3 \\
Without SFT & 25.3 & 21.7 \\
\bottomrule
\end{tabularx}

\caption{Complete 8B component ablations. Drops are relative to the 47.0 full setting.}
\label{tab:ablation}\end{table}
\begin{table}[htbp]\centering\small
\begin{tabularx}{\linewidth}{@{}Xrrr@{}}
\toprule
Scale & Direction & Magnitude & Magnitude share \\
\midrule
8B & 3.57 & 19.48 & 84.5\% \\
30B & 5.73 & 21.56 & 79.0\% \\
\bottomrule
\end{tabularx}

\caption{Symmetric decomposition of the product of displayed mean $D/M$ values. Contributions are points on the 0--100 score scale.}
\label{tab:decomposition}\end{table}
\begin{table}[htbp]\centering\small
\begin{tabularx}{\linewidth}{@{}Xrrrr@{}}
\toprule
Setting & D & M & S & Rank corr. \\
\midrule
Base 8B & 65.1 & 36.8 & 24.0 & 0.06 \\
Best global scale & 65.1 & 48.5 & 31.6 & 0.06 \\
Best scale per horizon & 65.1 & 52.5 & 34.2 & 0.11 \\
Trained, shuffled within horizon & 51.2 & 55.3 & 28.3 & 0.02 \\
Trained 8B & 72.1 & 65.2 & 47.0 & 0.31 \\
GLM-5.3 & 81.6 & 64.2 & 52.4 & 0.19 \\
\bottomrule
\end{tabularx}

\caption{Rescaling and within-horizon shuffling controls. Rank corr. measures magnitude rank correlation. Positive rescaling preserves predicted signs.}
\label{tab:controls}\end{table}
\begin{table}[htbp]\centering\small
\begin{tabularx}{\linewidth}{@{}Xrrr@{}}
\toprule
Metric & Base & SFT & RL \\
\midrule
Median tool calls & 4 & 19 & 17 \\
Individual-stock query share & 61\% & 38\% & 33\% \\
Technical-indicator query share & 19\% & 21\% & 18\% \\
Ranking / market-context share & 14\% & 27\% & 31\% \\
News query share & 6\% & 14\% & 18\% \\
Compactions per trajectory & 0.0 & 1.1 & 1.6 \\
\bottomrule
\end{tabularx}

\caption{Investigation behavior for the 8B base, SFT, and RL policies. Evidence-query shares and compaction counts are reported separately.}
\label{tab:behavior}\end{table}
\begin{table}[htbp]\centering\small
\begin{tabularx}{\linewidth}{@{}Xrrrrr@{}}
\toprule
Model & Short & Long & Easy & Medium & Hard \\
\midrule
Trained 8B & 49.3 & 43.7 & 59.8 & 47.1 & 29.8 \\
Trained 30B & 56.4 & 52.3 & 67.3 & 55.2 & 37.1 \\
GLM-5.3 & 48.9 & 57.4 & 70.9 & 52.6 & 27.4 \\
\bottomrule
\end{tabularx}

\caption{Scores by horizon and difficulty. Short/long counts are 232/164; easy/medium/hard counts are 136/158/102.}
\label{tab:subgroups}\end{table}

\FloatBarrier
\section{Reconstructing Context After Repeated Compaction}
A single investigation can compress its history more than once. For example, after checking a stock against its sector, the agent can compress the underlying price tables while retaining the dates and the relative move. If a later news query yields conflicting evidence, the agent can keep that article in full while compressing older indicator outputs.

Each compaction affects only subsequent decisions. To evaluate the probability of an earlier action during PPO, the update uses the context available when the agent chose that action. Replacing that context with a later summary would change the input to the policy and the critic.

\end{document}